\documentclass[11pt]{article}
\usepackage{acl}

\usepackage{times}
\usepackage{latexsym}
\usepackage[T1]{fontenc}
\usepackage[utf8]{inputenc}
\usepackage{microtype}
\usepackage{inconsolata}
\usepackage{graphicx}
\usepackage{booktabs}
\usepackage{multirow}
\usepackage{amsmath}
\usepackage{enumitem}
\usepackage{xspace}
\usepackage{tabularx}
\usepackage{array}
\usepackage{subcaption}
\usepackage{pifont}
\usepackage[most]{tcolorbox}

\newtcolorbox{qualexample}[1][]{
    enhanced,
    nobeforeafter,
    colback=blue!3,
    colframe=blue!50!black,
    boxrule=0.5pt,
    arc=2pt,
    left=6pt, right=6pt, top=5pt, bottom=5pt,
    title=#1,
    fonttitle=\bfseries\footnotesize,
    coltitle=white,
    colbacktitle=blue!50!black,
    attach boxed title to top left={xshift=6pt, yshift=-3pt},
    boxed title style={arc=2pt, boxrule=0pt},
    fontupper=\small,
}
\newcommand{\qfield}[2]{\noindent\textbf{#1:}~#2\par\vspace{1pt}}

\newcommand{\risk}{\textsc{risk suppression}\xspace}
\newcommand{\chair}{\textsc{CHAIR}\xspace}
\newcommand{\pope}{\textsc{POPE}\xspace}
\newcommand{\amber}{\textsc{AMBER}\xspace}
\newcommand{\mmstar}{\textsc{MMStar}\xspace}
\newcommand{\llavaonefive}{LLaVA-1.5\xspace}
\newcommand{\llavanext}{LLaVA-NeXT\xspace}
\newcommand{\instructblip}{InstructBLIP\xspace}

\title{Does Playing it Safe Count as Faithfulness? Reassessing LVLM Hallucination Mitigation Methods}

\author{Mehrdad Fazli\thanks{\ \ Equal contribution.}, Sina Mansouri\footnotemark[1], Mohit Marvania\footnotemark[1], Ziwei Zhu \\
Department of Computer Science, George Mason University \\
Fairfax, VA 22030, USA \\
\texttt{\{mfazli, smansou3, mmarvani, zzhu20\}@gmu.edu}}

\begin{document}
\maketitle
\begin{abstract}
Recent inference-time hallucination mitigation methods for large vision-language models (LVLMs) report strong gains on hallucination benchmarks. However, it remains unclear whether lower hallucination scores reflect improved multimodal grounding or more conservative generation. We evaluate six mitigation methods across three LVLMs and four benchmarks, including hallucination-focused evaluation and the diverse capability benchmark \mmstar. Our analysis reveals two consistent patterns. First, hallucination reduction is often coupled with reduced informativeness: methods that lower hallucination rates also reduce object recall, visual coverage, or response detailedness. Second, improvements on hallucination benchmarks do not reliably transfer to broader multimodal capabilities, with methods showing inconsistent or degraded performance on fine-grained perception and reasoning tasks. Our findings suggest that current evaluation protocols may overestimate progress by rewarding conservative generation. We argue that hallucination mitigation should be evaluated as a faithfulness--informativeness--capability trade-off rather than through hallucination scores alone. 
Data and code are available
at~\url{https://github.com/mehrdadfazli/AssessHalVLM}
\end{abstract}

\section{Introduction}

Large vision-language models (LVLMs) achieve strong performance across multimodal tasks but remain prone to visual hallucination, including \emph{object hallucination}: mentioning objects that are unsupported by the image. To address this, a growing body of work proposes training-free, decoding-time mitigation methods that adjust the model's output distribution without retraining \citep{vcd, opera, m3id, avisc}. These methods report substantial gains on standard hallucination benchmarks such as \chair~\cite{chair}, \pope~\cite{pope}, and \amber~\cite{amber}, and have become standard baselines for inference-time hallucination mitigation.

\begin{table*}[t]
\centering
\footnotesize
% \resizebox{\textwidth}{!}{%
\begin{tabular}{lll}
\hline
\textbf{Evaluation Axis} & \textbf{Component Class} & \textbf{Target Instantiations / Metrics} \\ \hline 
\textbf{Base Models} & Open-Source LVLMs & \llavaonefive, \llavanext, \instructblip \\ \hline
\multirow{3}{*}{\textbf{Mitigation Methods}} & Contrastive Decoding & VCD~\cite{vcd}, M3ID~\cite{m3id} \\
 & Attention Calibration & AGLA~\cite{agla}, CAAC~\cite{fazli2026caac} \\
 & Hidden-State Modification & CEI~\cite{cei}, AFTER~\cite{wang_after_2026} \\ \hline
{\multirow{2}{*}{\textbf{Hallucination Benchmarks}}} & \multirow{2}{*}{faithfulness-informativeness} & \chair~\cite{chair}\\
& & \amber~\cite{amber} \\ \hline
\textbf{Capability Benchmark} & vision-indispensable benchmark & \mmstar~\cite{mmstar} \\ \hline
\end{tabular}
% }
\caption{Systematic diagnostic evaluation matrix spanning our 54  experimental configurations, decoupling error mitigation from information retention and broad reasoning capability.}
\label{tab:evaluation_matrix}
\vspace{-10pt}
\end{table*}

We argue that these gains may be misleading. 
Many widely used hallucination benchmarks emphasize object-level hallucination in COCO-style natural images, where a method can lower hallucination scores by generating shorter, non-committal responses that mention fewer visual entities, minimizing opportunities to err without improving visual grounding. We call this behavior \emph{risk suppression}: reducing unsupported visual claims by reducing visual commitment, rather than by improving multimodal grounding.  Additionally, a method may score well on hallucination benchmarks while failing to improve, or even compromising, broader multimodal capabilities such as reasoning, and fine-grained perception. We call this disconnect the \emph{faithfulness--capability gap}: improvements in narrow faithfulness metrics do not reliably transfer to broader multimodal competence.

We conduct a diagnostic evaluation of six inference-time mitigation methods across three LVLMs and three benchmarks. We evaluate risk suppression on \chair and \amber, and faithfulness--capability gap on \mmstar to test whether mitigation gains transfer to perception, reasoning, science, and mathematics tasks. To detect risk suppression, we pair hallucination metrics with informativeness metrics drawn from the same benchmark: object recall for \chair and visual coverage (Cover) for \amber. Our analysis yields two findings:
\begin{itemize}
    \item \textbf{Hallucination reduction often reflects risk suppression.} Across both \chair and \amber, methods that reduce hallucination tend to reduce informativeness in tandem: lower CHAIR correlates with lower object recall, and lower AMBER hallucination correlates with lower Cover. The methods that score best on hallucination are often the most conservative, rather than the most informative.
    \item \textbf{Hallucination gains do not reliably transfer to broader capabilities.} On \mmstar, mitigation methods show inconsistent or degraded performance, with effects that flip in direction across models and categories. Gains on object-hallucination benchmarks frequently fail to translate to fine-grained perception and reasoning tasks such as math, science, and logical reasoning.
\end{itemize}

These results suggest that current evaluation protocols may overestimate progress on LVLM faithfulness by rewarding conservative decoding behavior. We argue that hallucination mitigation should be evaluated as a \emph{faithfulness--informativeness--capability trade-off} rather than by hallucination scores alone. 

\section{Diagnostic Evaluation Setup}
\label{sec:exp_setup}

To rigorously expose the trade-offs of inference-time hallucination mitigation, we conduct an evaluation spanning 3 LVLMs $\times$ 6 methods $\times$ 3 benchmarks, totaling 54 distinct experimental configurations (Table~\ref{tab:evaluation_matrix}). 

\textbf{Models \& Mitigation Methods.}
We evaluate the 7B-parameter variants of three widely adopted LVLMs~\cite{wang_after_2026, fazli2026caac} in the hallucination literature: \llavaonefive~\citep{llava15}, \llavanext~\citep{llavanext}, and \instructblip~\citep{instructblip}. 
For each architecture, we benchmark the vanilla decoding baseline against six representative inference-time hallucination mitigation methods spanning the three dominant families: \textit{contrastive decoding} (VCD~\citep{vcd}, M3ID~\citep{m3id}), \textit{attention calibration} (AGLA~\citep{agla}, CAAC~\citep{fazli2026caac}), and \textit{hidden-state modification} (CEI~\citep{cei}, AFTER~\citep{wang_after_2026}). Furthermore, this selection deliberately balances foundational, well-respected approaches (VCD, M3ID, AGLA) with recent state-of-the-art advancements (CAAC, CEI, AFTER) in hallucination mitigation.

\textbf{Benchmark Selection \& Decoupled Metrics.}
To systematically decouple hallucination reduction from information loss and general intelligence degradation, we partition our evaluation suite into two distinct categories:

\textit{Hallucination-Informativeness Suites:} We select two established benchmarks that specifically provide paired metrics for both hallucination rate and informativeness. \chair~\citep{chair} evaluates object-existence errors in open-ended MS-COCO captioning; we report $\text{CHAIR}_s$/$\text{CHAIR}_i$ (lower is better) paired with \textit{Object Recall} (higher is better) as an informativeness proxy. \amber~\citep{amber} evaluates generative responses; we report its object hallucination rate metrics (CHAIR and Hal) alongside \textit{Cover}, the fraction of ground-truth visual entities successfully mentioned.

\textit{Broader Multimodal Capability Benchmark:} To evaluate how hallucination mitigation methods influence broader model capabilities, we leverage \mmstar~\citep{amber, mmstar}. Crucially, \mmstar distills multimodal examples from existing benchmarks by purging samples solvable via textual cues or shallow visual shortcuts, enabling a rigorous evaluation of six core multimodal capabilities: coarse perception, fine-grained perception, instance reasoning, logical reasoning, science and technology, and mathematics. We report overall and category-specific accuracy.

\section{Hallucination Reduction as Risk Suppression}
\label{sec:finding1}
Genuine improvements in multimodal faithfulness require minimizing hallucination rates without sacrificing visual information density. Under the \textit{risk suppression} hypothesis, however, localized drops in hallucination errors often reflect a reduction in detailedness and visual commitment. An optimal model should navigate toward the upper-left quadrant of the original model in a hallucination--informativeness space, simultaneously achieving low error and high informativeness.
As demonstrated below, current mitigation methods instead shift models along a risk-hedging diagonal.
% --- FIGURE 1: TWO-PANEL PLOT ---
\begin{figure}[t]
\centering
\begin{subfigure}{0.38\textwidth}
    \centering
    \includegraphics[width=\linewidth]{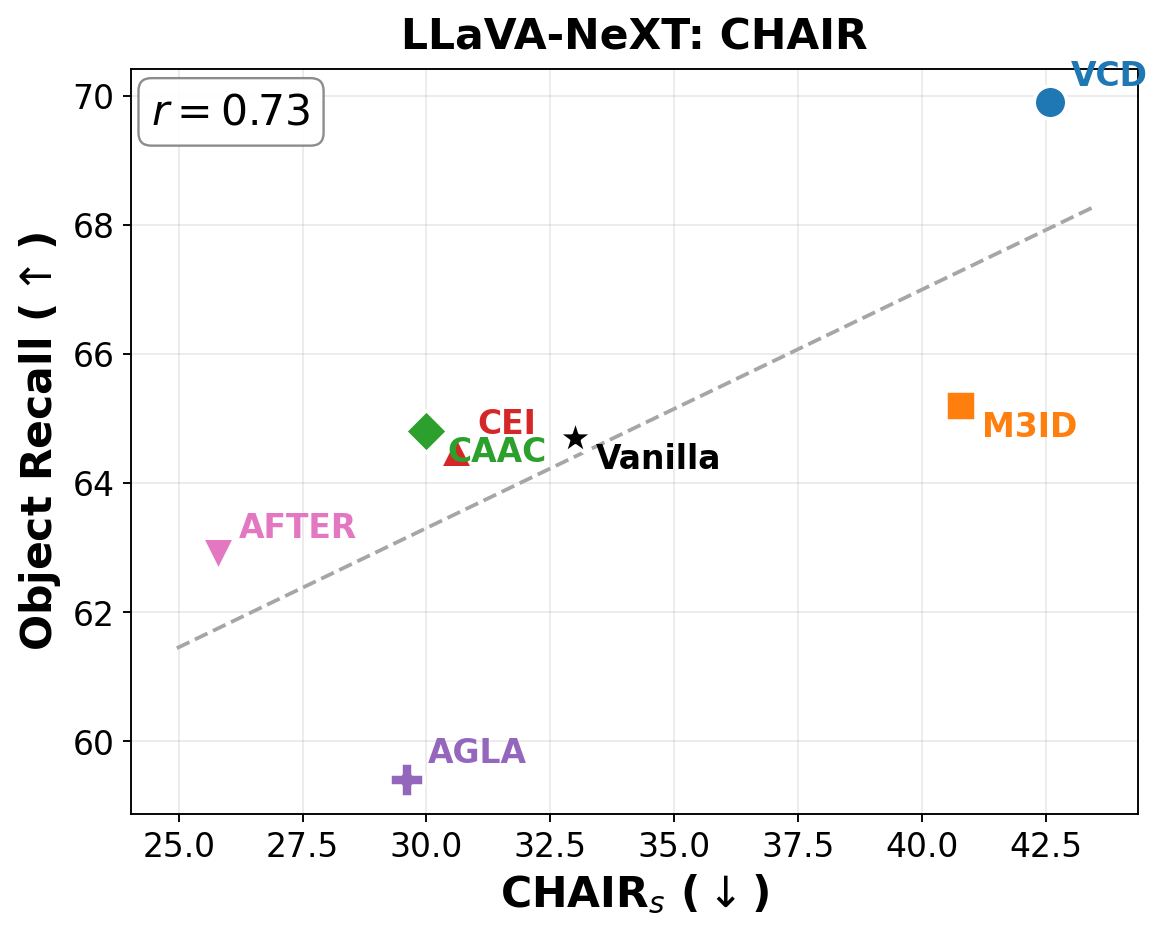}
    \caption{\chair: $\text{CHAIR}_s$ vs. Object Recall}
    \label{fig:chair_tradeoff}
\end{subfigure}
\begin{subfigure}{0.38\textwidth}
    \centering
    \includegraphics[width=\linewidth]{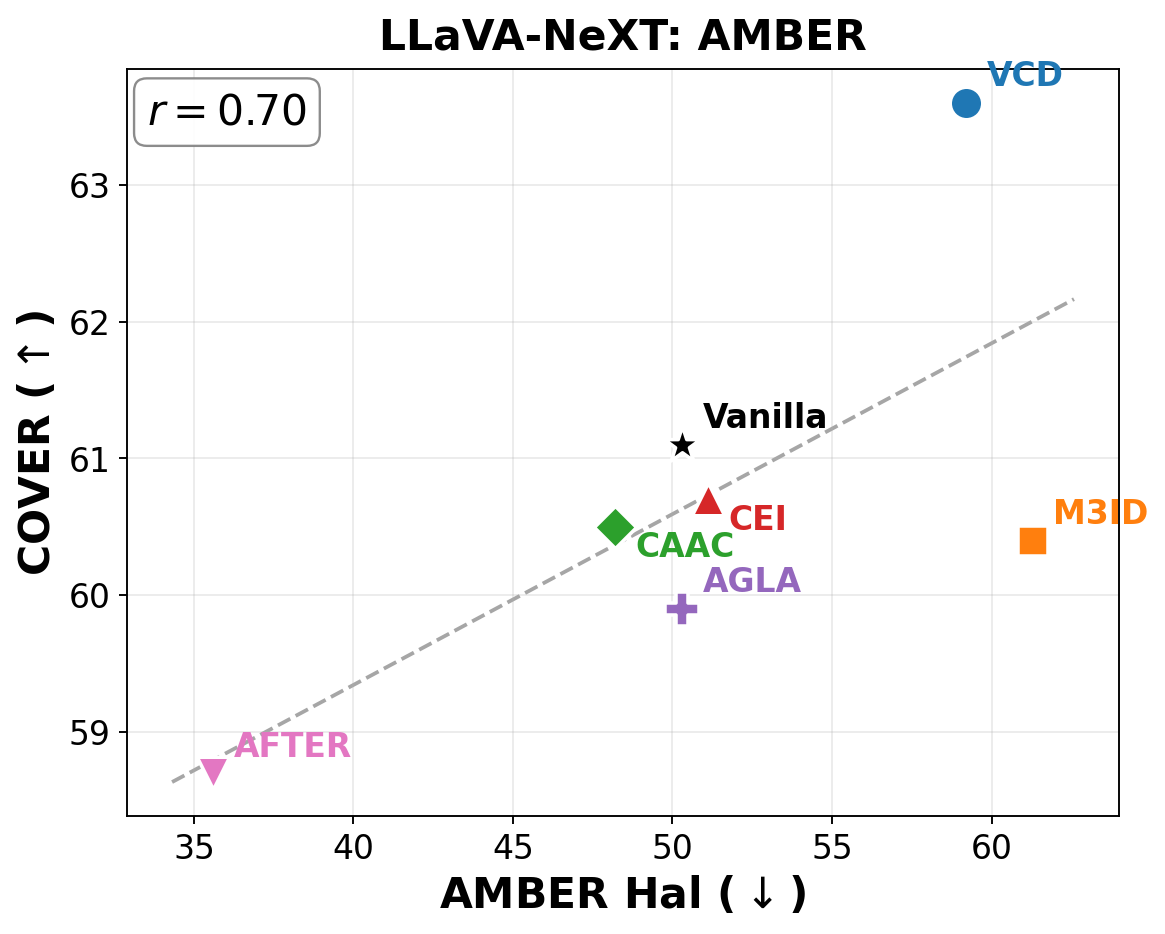}
    \caption{\amber: Hallucination Rate vs. Cover}
    \label{fig:amber_tradeoff}
\end{subfigure}
\vspace{-5pt}
\caption{The faithfulness--informativeness trade-off on \llavanext-7B. 
Both (a) \chair and (b) \amber exhibit a strict positive correlation between error rates and informativeness. 
The target optimal quadrant is the upper-left of the vanilla decoding (minimized errors, maximized recall).}

\label{fig:risk-suppression}
\vspace{-10pt}
\end{figure}

\begin{figure*}[t]
\centering
\includegraphics[width=0.85\textwidth]{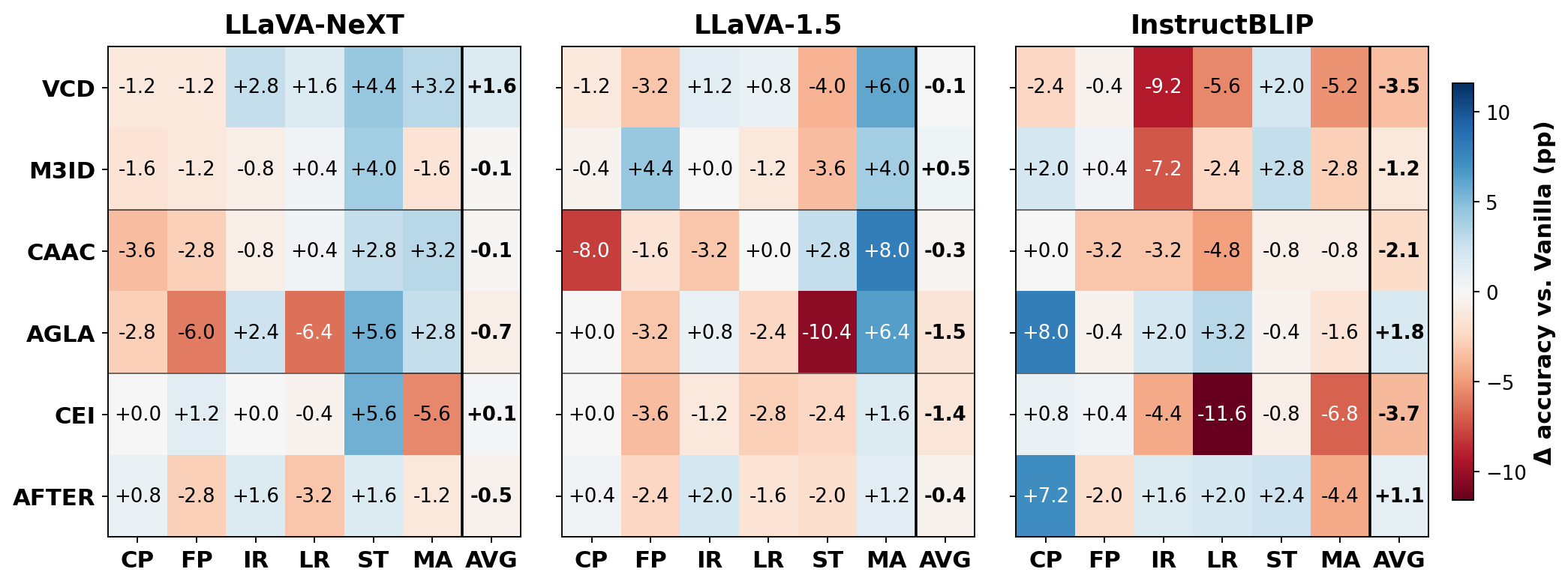}
\vspace{-5pt}
\caption{Per-category accuracy change ($\Delta$ vs. vanilla baseline, in percentage points) on \mmstar across three distinct LVLM architectures. Columns track specific \mmstar task categories: Coarse Perception (\texttt{CP}), Fine-grained Perception (\texttt{FP}), Instance Reasoning (\texttt{IR}), Logical Reasoning (\texttt{LR}), Science \& Technology (\texttt{ST}), and Mathematics (\texttt{MA}). Red cells indicate performance degradation, while blue cells mark performance improvements.}
\label{fig:mmstar-delta}
\vspace{-10pt}
\end{figure*}

Figure~\ref{fig:risk-suppression} illustrates these operational trade-offs for \llavanext (7B), with full cross-model distributions for \llavaonefive and \instructblip provided in Appendix~\ref{app:full-results}. 

\textbf{Information density trade-offs in captioning.} Across both evaluation suites, hallucination rates and informativeness exhibit a powerful positive correlation. 
On \chair, sentence-level hallucination ($\text{CHAIR}_s$) and object recall are strongly correlated (Pearson $r = 0.73$). Only CAAC and CEI preserved the recall while mitigating hallucination, whereas other methods like AFTER and AGLA dropped the recall as they reduced the hallucinations.
A similar pattern is manifested on \amber, where instance-level hallucination ($\text{Hal}$) and object coverage ($\text{Cover}$) correlate at $r = 0.70$. 
Here, AFTER achieves the lowest nominal hallucination rate but achieves this at the cost of reduced coverage. 
Interestingly, contrastive decoding frameworks (e.g., VCD and M3ID) actively amplify vision-conditioned tokens over language priors. This mechanism shifts models toward a state of hyper-loquacity—inducing an over-descriptiveness that simultaneously inflates both hallucination rates and informativeness. 
None of the evaluated methods could penetrate the upper-left quadrant and most followed the hallucination--informativeness diagonal.

\section{Hallucination Gains Do Not Reliably Transfer to General Capabilities}
\label{sec:finding2}

To verify if a decoding-time intervention genuinely improves visual grounding, we must evaluate whether it preserves or enhances broader visual understanding and reasoning capabilities. Under the \risk framework, we hypothesize that methods enforcing a more conservative, less detailed output distribution will suffer performance regressions on benchmarks requiring explicit, non-trivial visual commitments. We evaluate this premise using \mmstar~\citep{mmstar}, testing across six distinct multimodal categories where models cannot rely on simple textual cues or risk-averse generation to succeed.

\begin{figure}[t]
\centering
\includegraphics[width=\columnwidth]{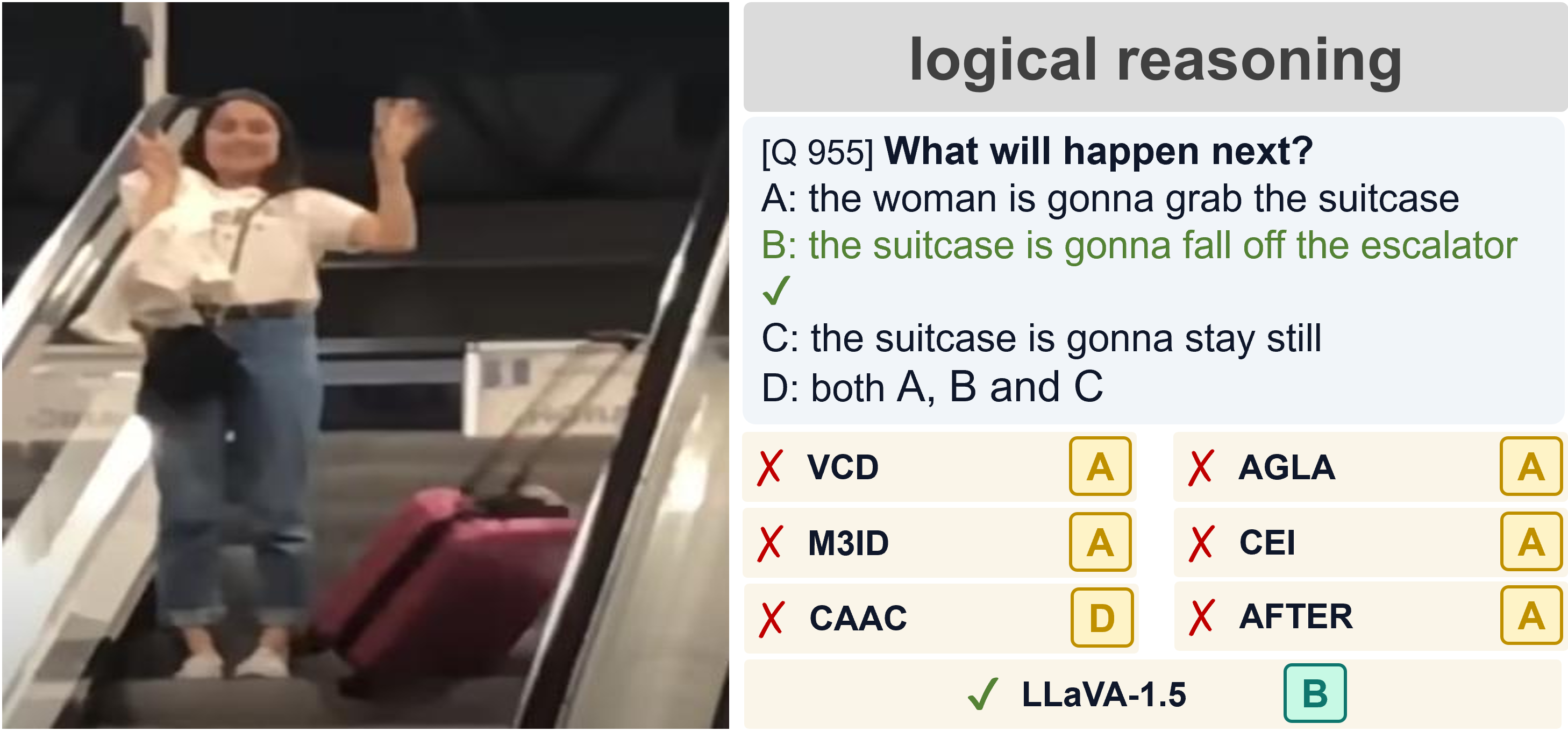}
\vspace{-15pt}
\caption{An \mmstar example demonstrating logical reasoning collapse. While the vanilla \llavaonefive accurately reasons about the image, all six mitigation methods lead to incorrect options.}
\label{fig:case_study_escalator}
\vspace{-10pt}
\end{figure}

Figure~\ref{fig:mmstar-delta} displays accuracy changes ($\Delta$) relative to vanilla decoding across all 54 configurations, across six multimodal tasks from \mmstar. 
Our matrix analysis reveals two key observations:

\textbf{Impairing fine-grained perception and reasoning tasks.}
decoding modifications systematically damage fine-grained perception (\texttt{FP}) across the board. 
Out of all 18 evaluated \texttt{FP} configurations, only two settings managed to yield isolated gains of more than 1\%, while the vast majority across all three models suffered performance degradation. Similarly, the mitigation methods overwhelmingly degrade the instance reasoning and logical reasoning particularly in \instructblip and \llavaonefive. Figure~\ref{fig:case_study_escalator} showcases an example in \texttt{LR} where every single mitigation strategy erodes the base model's accurate response. More examples are provided in Appendix~\ref{app:qualitative}.

\textbf{Most effects are modest and inconsistent.} No method consistently improves \mmstar over vanilla across models or categories; in fact, every tested mitigation method fails to improve macro averages on more than one out of the three base architectures. Additionally, gains in one category are typically offset by losses elsewhere.
For instance, AFTER boosts coarse perception (\texttt{CP}) by $7.2\%$ on \instructblip, yet simultaneously undermines mathematical reasoning (\texttt{MA}) by $4.4\%$ on the same architecture. 
Similarly, while AGLA on \llavaonefive yields a $6.4\%$ increase in mathematics, it compromises performance on scientific and technical tasks, dropping \textit{Science \& Technology} (\texttt{ST}) by a substantial $10.4$ percentage points.

\section{Related Work and Discussion}
\label{sec:related_work}

\paragraph{Inference-time hallucination mitigation.}
Training-free inference interventions have become a dominant paradigm for hallucination mitigation~\citep{bai_hallucination_2025, liu_survey_2024}. Contrastive decoding strategies bias token selection toward image-supported representations using alternative distributions~\citep{chuang_dola_2024, vcd, zhu_ibd_2024, chen_ict_2024}. Concurrently, attention-based methods intervene directly on visual grounding by amplifying image attention~\citep{liu_paying_2024}, calibrating attention patterns~\citep{zhu_mitigating_2025, zhang_seeing_2024}, or penalizing uninformative and over-trusted tokens~\citep{opera, avisc}.

\paragraph{Hallucination benchmarks.}
Hallucination benchmarks are commonly discriminative, such as POPE~\cite{pope}, MME~\cite{fu_mme_2024}, and HallusionBench~\cite{guan_hallusionbench_2024}, or generative, such as CHAIR~\cite{chair}, and AMBER~\cite{amber}.
Discriminative benchmarks provide objective yes/no-style evaluation, while generative benchmarks evaluate free-form responses but remain largely object-centric. 
Most hallucination mitigation studies primarily report hallucination rates, with limited evaluation of informativeness or broader multimodal capabilities. 
In contrast, our work jointly evaluates hallucination, informativeness, and transfer to general LVLM capability benchmarks.

\section{Conclusion}
\label{sec:conclusion}
% We presented a diagnostic study of six inference-time hallucination mitigation methods across three LVLMs and three benchmarks. Our results show that hallucination reductions on \chair and \amber often coincide with reduced informativeness, and that hallucination-benchmark gains do not reliably transfer to broader multimodal capabilities on \mmstar. These patterns are consistent with \risk: methods that score best on hallucination metrics often achieve this by curtailing the model's generation of specific visual claims rather than by clearly improving multimodal grounding. We argue that hallucination mitigation should be evaluated as a faithfulness--informativeness--capability trade-off, and that hallucination rate improvements alone are insufficient evidence of better multimodal grounding.

We presented a diagnostic study of six inference-time hallucination mitigation methods across three LVLMs and three benchmarks. Our results show that hallucination reductions on \chair and \amber often coincide with reduced informativeness, and these gains do not reliably transfer to broader multimodal capabilities on \mmstar. 
% This pattern is consistent with \risk: top-scoring methods on hallucination metrics often achieve this by curtailing specific visual claims rather than improving multimodal grounding. 
We argue that hallucination mitigation must be evaluated as a faithfulness--informativeness--capability trade-off, as lower hallucination rates alone provide insufficient evidence of better visual grounding.

\section*{Limitations}

\paragraph{Benchmark coverage.}
Our evaluation covers three benchmarks (\chair, \amber, \mmstar). While these span object-level hallucination, generative coverage, LLM-judge scoring, and broad multimodal capability, they do not cover all relevant axes of LVLM faithfulness, such as attribute-level or relational hallucination, video understanding, or long-form instruction following. The patterns reported here may not extend uniformly to those settings.

\paragraph{Scope of mitigation methods.}
We focus on representative inference-time hallucination mitigation methods, including decoding-time and attention-based interventions. These methods are widely used because they are training-free and can be applied to existing LVLMs, but they do not cover all possible mitigation strategies like training-based approaches, preference optimization, and retrieval-augmented grounding. Our conclusions therefore apply most directly to inference-time distribution-shaping methods.

\paragraph{Model scale.}
We evaluate three 7B-scale LVLMs. Larger or more recent models may respond differently to decoding-time interventions, and the trade-offs we observe may vary with model scale and architecture.

\paragraph{Hyperparameter Scoping.} We implement each mitigation method using the hyperparameter configurations specified by their respective authors. When applying a method to an LVLM not included in its original study, we maintain these default settings rather than conducting exhaustive per-model optimization. While custom tuning might alter individual performance profiles, our approach reflects standard, out-of-the-box deployment conditions.

% For each mitigation method, we use the hyperparameters reported in the original publication. Per-model tuning could improve individual results or shift a method along the hallucination--informativeness trade-off. We leave systematic hyperparameter sweeps to future work.

\paragraph{Causal interpretation.}
Our analysis demonstrates that hallucination metrics are entangled with informativeness and do not reliably transfer to broader capabilities. We do not directly measure ``grounding'' as a latent property; \risk describes the observed evaluation pattern, not a mechanistic claim about why each method behaves as it does.

\paragraph{Potential Risks.}
While our work primarily focuses on the critical analysis and evaluation of existing hallucination mitigation methods, a potential risk is that highlighting the systematic trade-offs of these techniques could inadvertently guide the development of generation strategies that look superficially accurate while masking underlying unreliability. Additionally, understanding how to bypass or exploit conservative decoding behaviors could theoretically be used to manipulate large vision-language models (LVLMs) in safety-critical applications.
% \paragraph{Judge dependence.}
% \mmhal results depend on the choice of LLM judge. We use GPT-4o with temperature~0 and fixed prompts, but absolute scores may vary with judge model, version, or prompt. Our \mmhal analysis emphasizes relative comparisons across methods within the same judge configuration.

\paragraph{AI Assistance Disclosure}
AI language models were utilized for text editing, grammatical refinement, and coding assistance. The authors have fully reviewed and verified all technical content, experimental results, and core claims.

\bibliography{references}

\appendix
\section{Full Experimental Details}
\label{app:experimental-details}

\subsection{Hyperparameters}
\label{app:hyperparams}

We organize hyperparameters by method family. Within each family we follow the values reported in the original publication; deviations are flagged explicitly. The full per-method, per-benchmark configuration is also provided in the released codebase at:~\url{https://anonymous.4open.science/r/AssessHalVLM-48FB/}.

\paragraph{Contrastive decoding methods.}
VCD \citep{vcd} and M3ID \citep{m3id} both compute a contrast between the model's logits on the original image and its logits on a perturbed logit distribution, then subtract the reference's distribution from the conditional one to lower the hallucinatory token's probabilities. Both methods retain only tokens that exceed an adaptive plausibility threshold (controlled by $\beta$) to avoid amplifying tail noise. Table~\ref{tab:hyper-cd} lists the values used for both methods.

% \begin{table*}[t]
% \centering
% \small
% \begin{tabular}{lll}
% \toprule
% Method & Parameter & Value \\
% \midrule
% VCD \citep{vcd}   & $\alpha$ (contrast strength)         & 1.0 \\
%                   & $\beta$ (plausibility threshold)     & 0.1 \\
%                   & DDPM noise step                      & 500 \\
% \midrule
% M3ID \citep{m3id} & $\lambda$ (amplification coefficient) & 0.2 \\
%                   & $\beta$ (plausibility threshold)      & 0.1 \\
% \bottomrule
% \end{tabular}
% \caption{Hyperparameters for contrastive-decoding methods. The M3ID $\lambda{=}0.2$ matches the runner-level default of the CAAC baselines codebase \citep{fazli2026caac}; the original M3ID paper reports a slightly different default that we did not retune for fairness across cells.}
% \label{tab:hyper-cd}
% \end{table*}
\begin{table}[!htbp]
\centering
\footnotesize
\setlength{\tabcolsep}{4pt}
\renewcommand{\arraystretch}{1.2}
\begin{tabular}{@{}p{1.6cm}p{3.0cm}p{1.2cm}@{}}
\toprule
\textbf{Method} & \textbf{Parameter} & \textbf{Value} \\
\midrule
\multirow{3}{1.6cm}{VCD \citep{vcd}}
                  & $\alpha$ (contrast strength)     & 1.0 \\
                  & $\beta$ (plausibility threshold) & 0.1 \\
                  & DDPM noise step                  & 500 \\
\midrule
\multirow{2}{1.6cm}{M3ID \citep{m3id}}
                  & $\lambda$ (amplification coef.)  & 0.2 \\
                  & $\beta$ (plausibility threshold) & 0.1 \\
\bottomrule
\end{tabular}
\caption{Hyperparameters for contrastive-decoding methods. The M3ID $\lambda{=}0.2$ matches the runner-level default of the CAAC baselines codebase \citep{fazli2026caac}; the original M3ID paper reports a slightly different default that we did not retune for fairness across cells.}
\label{tab:hyper-cd}
\end{table}

\paragraph{Attention-based methods.}
CAAC \citep{fazli2026caac} calibrates the attention weights placed on visual tokens at decoding time, downweighting tokens whose attention concentration is inconsistent with overall visual saliency. AGLA \citep{agla} assembles two attention paths: a \emph{global} path attending to the full image and a \emph{local} path attending to a masked, augmented image; the per-token logits are mixed additively as $\mathrm{logits} + \alpha \cdot \mathrm{logits}_{\mathrm{cd}}$, with an adaptive masking ratio derived from a BLIP image--text-matching score \citep{li2022blip}. We use the released default configurations for both methods, summarised in Table~\ref{tab:hyper-attn}.

\begin{table}[!htbp]
\centering
\footnotesize
\setlength{\tabcolsep}{3pt}
\renewcommand{\arraystretch}{1.15}
\begin{tabular}{@{}p{1.0cm}p{2.0cm}p{3.5cm}@{}}
\toprule
\textbf{Method} & \textbf{Parameter} & \textbf{Value} \\
\midrule
\multirow{3}{1.0cm}{CAAC \citep{fazli2026caac}}
                   & Calibration scope              & All decoder layers \\
                   & Confidence aggregation         & Token-level (default) \\
                   & Inference budget               & Matches vanilla \\
\midrule
\multirow{5}{1.0cm}{AGLA \citep{agla}}
                   & $\alpha$ (mixing ratio)        & 2.0 \\
                   & $\beta$ (plausibility threshold) & 0.5 \\
                   & Augmentation backbone          & BLIP-ITM (large), GradCAM block 6 \\
                   & Masking ratio                  & $1 - \mathrm{ITC}/2$ \\
                   & Inference budget               & Matches vanilla \\
\bottomrule
\end{tabular}
\caption{Hyperparameters for attention-based methods. Both methods are applied with the unmodified released configurations from their official codebases. For \llavanext, AGLA's augmented images are precomputed in a separate environment with \texttt{transformers}~4.34 (required by LAVIS-BLIP) and consumed by a token-by-token contrastive-decoding loop in \texttt{transformers}~4.47.}
\label{tab:hyper-attn}
\end{table}

% \paragraph{Hidden-state methods.}
% CEI \citep{cei} injects a contextual embedding into a fixed decoder layer at inference time, scaled by an injection strength $\gamma$. AFTER \citep{after} transfers attention activations from a designated source pattern (image+query) onto a target pattern (text+query) at selected layers, then continues decoding from the modified state. Both methods leave model parameters untouched. Configurations are summarised in Table~\ref{tab:hyper-hs}.

% \begin{table*}[t]
% \centering
% \small
% \begin{tabular}{lll}
% \toprule
% Method & Parameter & Value \\
% \midrule
% CEI \citep{cei}    & injection layer                      & released default \\
%                    & injection strength $\gamma$          & released default \\
%                    & decoded length budgets               & 64 / 128 / 256 / 512 \\
% \midrule
% AFTER \citep{after}& source attention                     & Image$+$Query \\
%                    & target attention                     & Text$+$Query \\
%                    & layer span                           & 64 / 7 (released best config) \\
%                    & decoded length budgets               & 64 / 512 \\
% \bottomrule
% \end{tabular}
% \caption{Hyperparameters for hidden-state methods. For CEI and CAAC we report results at the 512-token budget (the longest produced by both methods) to match the open-ended evaluation protocol used for \chair and \amber.}
% \label{tab:hyper-hs}
% \end{table*}

\paragraph{Hidden-state methods.}
CEI \citep{cei} injects a contextual embedding into a fixed decoder layer at inference time, with the injection strength scheduled by a cosine over Top-$K$ probability-mass statistics. AFTER \citep{wang_after_2026} transfers attention activations from a source pattern (image+query) onto a target pattern (text+query) at the top-$K$ heads selected from AMBER training activations, then continues decoding from the modified state. Both methods leave model parameters untouched.

The architecture-level hyperparameters used for both methods are summarized in Table~\ref{tab:hyper-hs}. CEI uses per-model values released by its authors: $(\alpha_{\max}, \beta, K_{\mathrm{mass}}) = (0.40, 0.70, 40)$ for \instructblip, $(0.25, 0.55, 40)$ for \llavaonefive, and $(0.17, 0.35, 80)$ for \llavanext, with injection always at decoder layer 10. AFTER uses $K{=}64$ heads with intervention strength $\alpha{=}7$, applied at the last token, with the QAO offset MLP pre-trained on AMBER for 10 epochs at learning rate $10^{-3}$.

% \begin{table}[t]
% \centering
% \small
% \setlength{\tabcolsep}{6pt}
% \renewcommand{\arraystretch}{1.2}
% \begin{tabular}{lll}
% \toprule
% Method & Parameter & Value \\
% \midrule
% CEI \citep{cei}    & Injection layer                      & 10 \\
%                    & Schedule                              & Cosine on Top-$K$ mass \\
%                    & Word-start gating $\delta$            & 0.5 \\
%                    & Per-model scalars                     & See prose above \\
% \midrule
% AFTER \citep{wang_after_2026} & Top-$K$ heads              & 64 \\
%                    & Intervention strength $\alpha$        & 7 \\
%                    & Edit location                         & Last token \\
%                    & Source $\to$ target attention         & Image$+$Q $\to$ Text$+$Q \\
% \bottomrule
% \end{tabular}
% \caption{Architecture-level hyperparameters for hidden-state methods. CEI's per-model scalars $(\alpha_{\max}, \beta, K_{\mathrm{mass}})$ are listed in the preceding paragraph since they vary by LVLM. Decoding-length budgets are not method hyperparameters and are pooled in the Decoding Settings section below.}
% \label{tab:hyper-hs}
% \end{table}
\begin{table}[!htbp]
\centering
\footnotesize
\setlength{\tabcolsep}{3pt}
\renewcommand{\arraystretch}{1.15}
\begin{tabular}{@{}p{1.0cm}p{2.2cm}p{3.3cm}@{}}
\toprule
\textbf{Method} & \textbf{Parameter} & \textbf{Value} \\
\midrule
\multirow{4}{1.0cm}{CEI \citep{cei}}
                   & Injection layer              & 10 \\
                   & Schedule                     & Cosine on Top-$K$ mass \\
                   & Word-start gating $\delta$   & 0.5 \\
                   & Per-model scalars            & See prose above \\
\midrule
\multirow{4}{1.0cm}{AFTER \citep{wang_after_2026}}
                   & Top-$K$ heads                & 64 \\
                   & Intervention $\alpha$        & 7 \\
                   & Edit location                & Last token \\
                   & Source $\to$ target          & Image$+$Q $\to$ Text$+$Q \\
\bottomrule
\end{tabular}
\caption{Architecture-level hyperparameters for hidden-state methods. CEI's per-model scalars $(\alpha_{\max}, \beta, K_{\mathrm{mass}})$ are listed in the preceding paragraph since they vary by LVLM. Decoding-length budgets are pooled in the Decoding Settings section below.}
\label{tab:hyper-hs}
\end{table}

\paragraph{Vanilla baseline.}
The vanilla decoding baseline for each model utilizes its standard, native generation configuration stripped of external modifications—meaning no contrastive terms, no attention reweighting, and no hidden-state calibration. Crucially, to eliminate stochastic decoding variability and isolate the exact impacts of each inference intervention, the vanilla baseline is uniformly restricted to deterministic greedy decoding across all evaluation suites, strictly mirroring the prompt constraints and token length budgets enforced during mitigation testing.

\paragraph{Implementation notes.}
VCD and M3ID are implemented from the CAAC baselines codebase, with patches limited to (i) adding \llavanext model-loading support and (ii) filtering unsupported sampling kwargs in the decoding utilities; the method logic is unmodified. AGLA is implemented from its official repository; we adapt the original POPE evaluation scripts to \chair, \amber, and \mmstar by removing the ``answer with one word'' prompt suffix.  The evaluation benchmarks (e.g., MMStar) and LVLM weights used in this study are publicly available under their respective open-source licenses
\subsection{Decoding Settings}

All runs use \texttt{seed=42} (Python, NumPy, PyTorch) with greedy decoding for the vanilla baseline. The \texttt{max\_new\_tokens} budget is $512$ for \chair and \amber and $32$ for \mmstar. Minor per-codebase deviations (e.g., the AGLA repo's default of multinomial sampling at $T{=}1.0$ for its target models) are documented in the released code.

\subsection{Benchmark Descriptions}
\label{app:benchmarks}

We use three benchmarks chosen to separate object-level hallucination from broader visual informativeness and from general multimodal capability. The hallucination-focused benchmarks (\chair, \amber) each pair an error metric with an informativeness metric drawn from the \emph{same} response; the capability benchmark (\mmstar) covers reasoning, science, and mathematics categories that hallucination scores do not measure.

\paragraph{\chair \citep{chair}.}
The Caption Hallucination Assessment with Image Relevance (CHAIR) protocol measures object-existence errors in open-ended captioning on MSCOCO images. Given a generated caption, an automatic pipeline extracts mentioned object nouns, compares them against MSCOCO's ground-truth object annotations for the same image, and reports two complementary rates: $\text{CHAIR}_s$ (the fraction of \emph{sentences} containing at least one hallucinated object) and $\text{CHAIR}_i$ (the fraction of all mentioned \emph{instances} that are hallucinated). We pair these error metrics with \emph{object recall}, the fraction of ground-truth objects that the caption successfully mentions, which serves as an informativeness proxy. We evaluate on 500 images sampled with \texttt{seed=42} from the COCO 2014 validation split; the generation prompt is ``\textit{Please describe this image in detail.}''.

\paragraph{\amber \citep{amber}.}
The Adversarial Multimodal Benchmark for hallucination Evaluation in LVLMs uses a curated set of 1004 generative items spanning object existence, attributes, and inter-object relations, with dense human-annotated object lists per image. We use the full generative split with prompt ``\textit{Describe this image.}'' AMBER reports two error metrics over the generated response --- $\text{CHAIR}$ (object-existence hallucination) and $\text{Hal}$ (instance-level hallucination rate including attribute and relation errors) --- and one informativeness metric, $\text{Cover}$, the fraction of the image's ground-truth objects mentioned. Throughout the paper we use $\text{Hal}$ paired with $\text{Cover}$, since this pairing is the most directly comparable to CHAIR's $(\text{CHAIR}_s, \text{Recall})$.

% \paragraph{\mmhal \citep{mmhal}.}
% MMHal-Bench evaluates LVLM responses with an LLM judge rather than a fixed object lexicon. It contains 96 image--question pairs across eight question types (object existence, counting, attributes, comparison, spatial relations, environment, holistic description, and adversarial), each constructed so that natural-image priors make a confidently wrong answer plausible. We prompt the model with the bare question (no MCQ suffix). Responses are scored 0--6 by a GPT-4o judge (model string \texttt{openai/gpt-4o} via the OpenRouter API, temperature 0.0, fixed system and user prompts taken verbatim from the official MMHal-Bench release. We report two summary statistics: the \emph{average score} (higher is better), and the \emph{hallucination rate}, defined following prior work as the fraction of responses scored strictly below 3 (lower is better).

\paragraph{\mmstar \citep{mmstar}.}
\mmstar is a vision-indispensable multimodal benchmark constructed by filtering existing evaluation suites to remove items that are solvable from text priors alone. The result is 1500 carefully audited multiple-choice questions (250 per category) covering six capability axes: \emph{coarse perception}, \emph{fine-grained perception}, \emph{instance reasoning}, \emph{logical reasoning}, \emph{science \& technology}, and \emph{mathematics}. Each item presents the image and a four-option MCQ; we append the VLMEvalKit suffix ``\textit{Please select the correct answer from the options above.}'' to elicit a single-letter response. We report per-category accuracy and the macro average across the six categories. \mmstar serves as our general-capability probe: a mitigation method that genuinely improves multimodal grounding should at least preserve performance on these reasoning-heavy tasks, not just on object-existence benchmarks like \chair or \amber.

% \paragraph{Why these three together.}
% The two hallucination benchmarks differ in the shape of the response they elicit (free captions, instructed descriptions) and in how hallucination is measured (object lexicon, hybrid lexicon); covering all three guards against benchmark-specific artifacts. Pairing each error metric with the corresponding informativeness metric (Recall or Cover) is what lets us read the risk-suppression pattern off the same response. \mmstar then provides an independent, capability-based check that no longer involves object-existence judgements at all.

\subsection{Hardware and Software}

All generation runs use a single NVIDIA A100-SXM4-80GB GPU. Judge runs are CPU-only. \llavaonefive and \instructblip use PyTorch 2.0.1, and \texttt{transformers} 4.31. \llavanext requires \texttt{transformers} 4.47 and uses PyTorch 2.5.1 with bitsandbytes 0.45 for 8-bit quantization. HuggingFace checkpoints used for the three LVLMs are: \texttt{llava-hf/llava-v1.6-vicuna-7b-hf}, \texttt{llava-hf/llava-1.5-7b-hf}, \texttt{Salesforce/instructblip-vicuna-7b}
% \textcolor{blue}{}All generation runs use a single NVIDIA A100-SXM4-80GB GPU ; judge runs are CPU-only. \llavaonefive and \instructblip run on PyTorch 2.0.1 with \texttt{transformers}~4.31; \llavanext requires \texttt{transformers}~4.47 with PyTorch 2.5.1 and bitsandbytes 0.45 for 8-bit quantization. CAAC-baseline runs load models with \texttt{load\_in\_8bit=True}; peak memory is $\sim$48~GB for AGLA (target LVLM + co-resident BLIP-ITM) and $\sim$20--25~GB otherwise. HuggingFace checkpoints: \texttt{llava-hf/llava-v1.6-vicuna-7b-hf}, \texttt{llava-hf/llava-1.5-7b-hf}, \texttt{Salesforce/instructblip-vicuna-7b}; AGLA additionally uses \texttt{liuhaotian/llava-v1.5-7b} and the LAVIS \texttt{blip2\_vicuna\_instruct} and \texttt{blip\_image\_text\_matching} (large) models.

% \subsection{Judge Prompt}
% \label{app:judge-prompt}

% The MMHal-Bench judge prompt is the original prompt from \citet{mmhal}, reproduced verbatim. It contains a task description, five worked examples spanning correct, incorrect, and ambiguous LMM responses, and a rating rubric (scores 0--6 jointly encoding informativeness and hallucination presence). The full prompt is approximately 4\,KB; we refer to the supplementary materials and to the original MMHal-Bench codebase\footnote{\url{https://huggingface.co/datasets/Shengcao1006/MMHal-Bench}}.

\section{Full Results}
\label{app:full-results}

This appendix reports the complete numerical results, per-model scatter plots complementing Figure~\ref{fig:risk-suppression}, an alternative bar-plot view of \mmstar deltas, and full \chair, \amber, and \mmstar result tables.

\subsection{Per-Model Trade-off Scatters}
\begin{figure}[p]
\centering
\begin{subfigure}{0.38\textwidth}
    \centering
    \includegraphics[width=\linewidth]{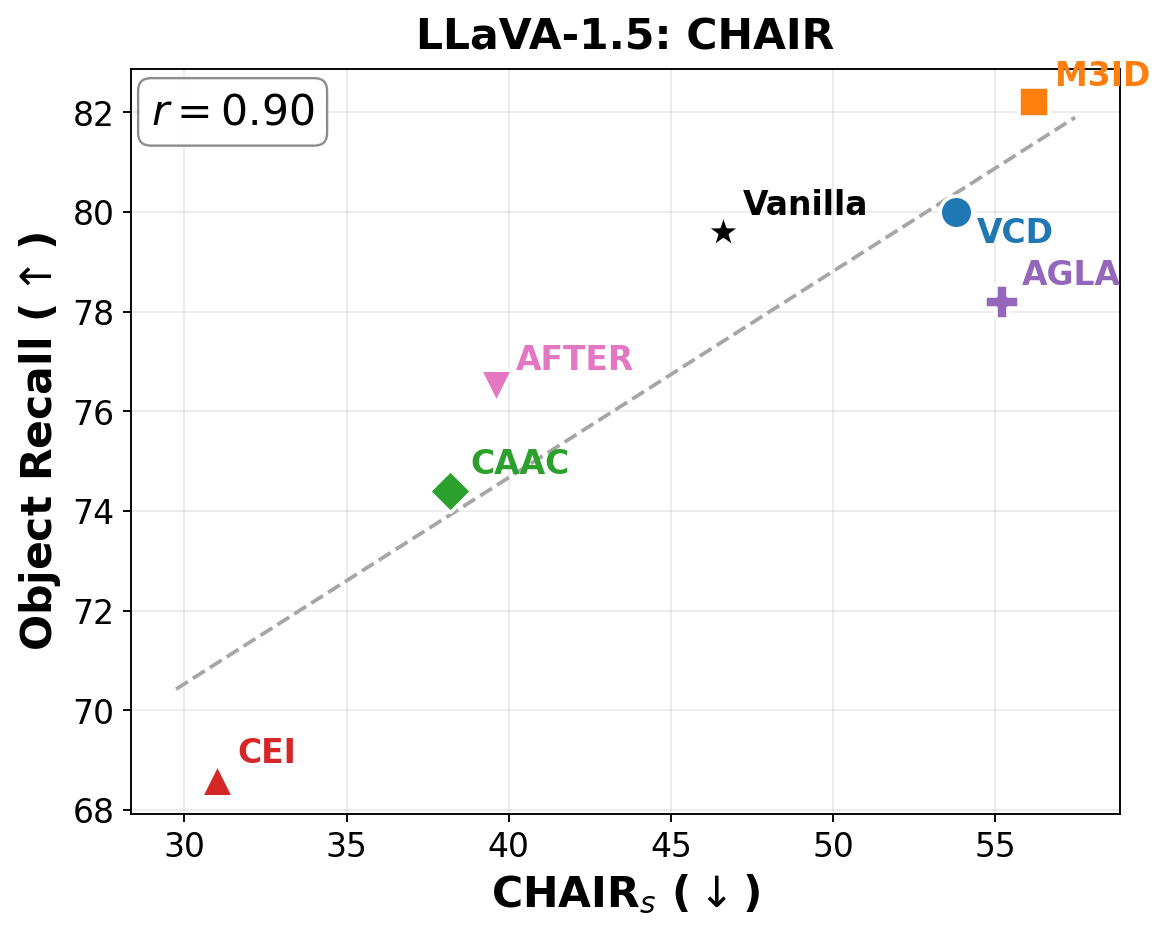}
    \caption{\chair: CHAIRs vs Recall}
    \label{fig:chair-llava15}
\end{subfigure}
\begin{subfigure}{0.38\textwidth}
    \centering
    \includegraphics[width=\linewidth]{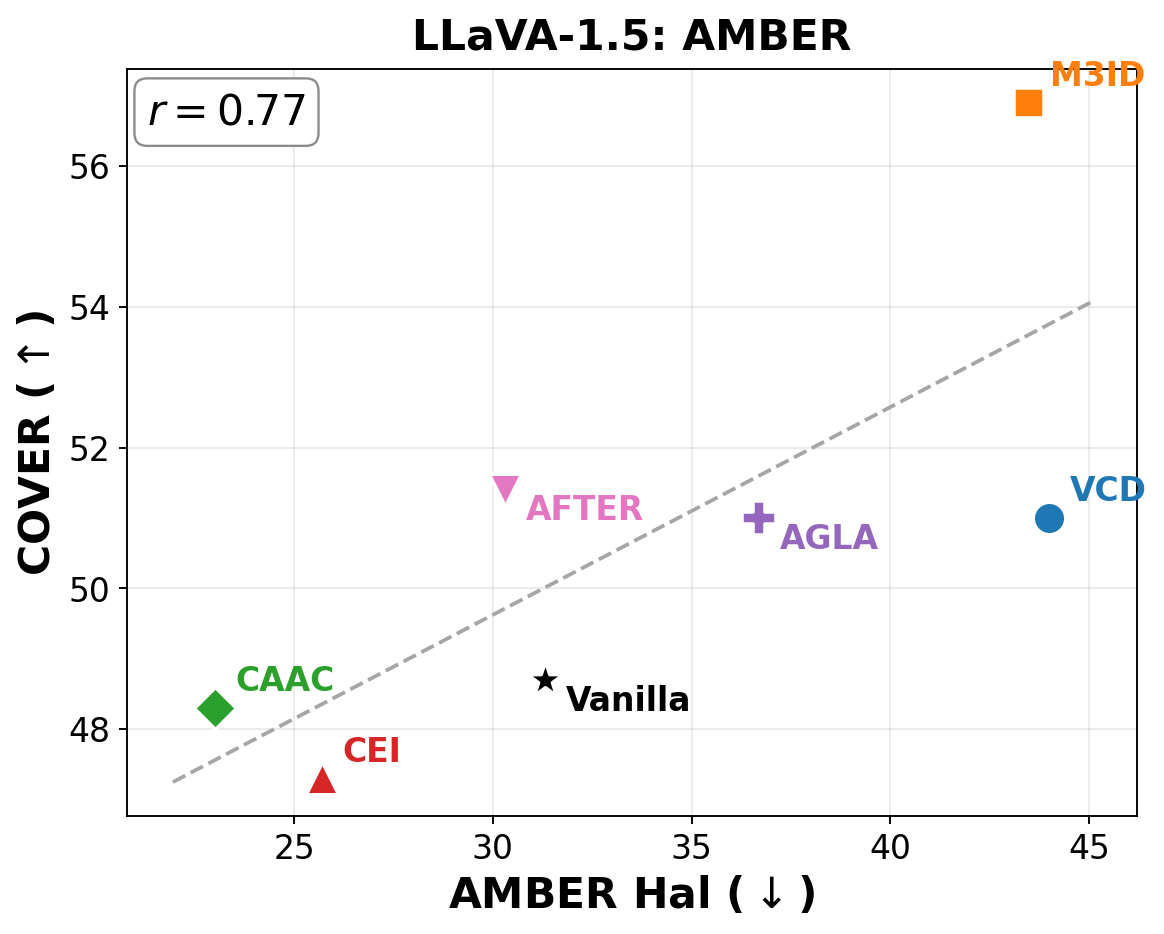}
    \caption{\amber: Hal vs Cover}
    \label{fig:amber-llava15}
\end{subfigure}
\caption{Hallucination--informativeness trade-off on \llavaonefive.}
\label{fig:appendix-llava15-scatter}
\end{figure}

\begin{figure}[p]
\centering
\begin{subfigure}{0.38\textwidth}
    \centering
    \includegraphics[width=\linewidth]{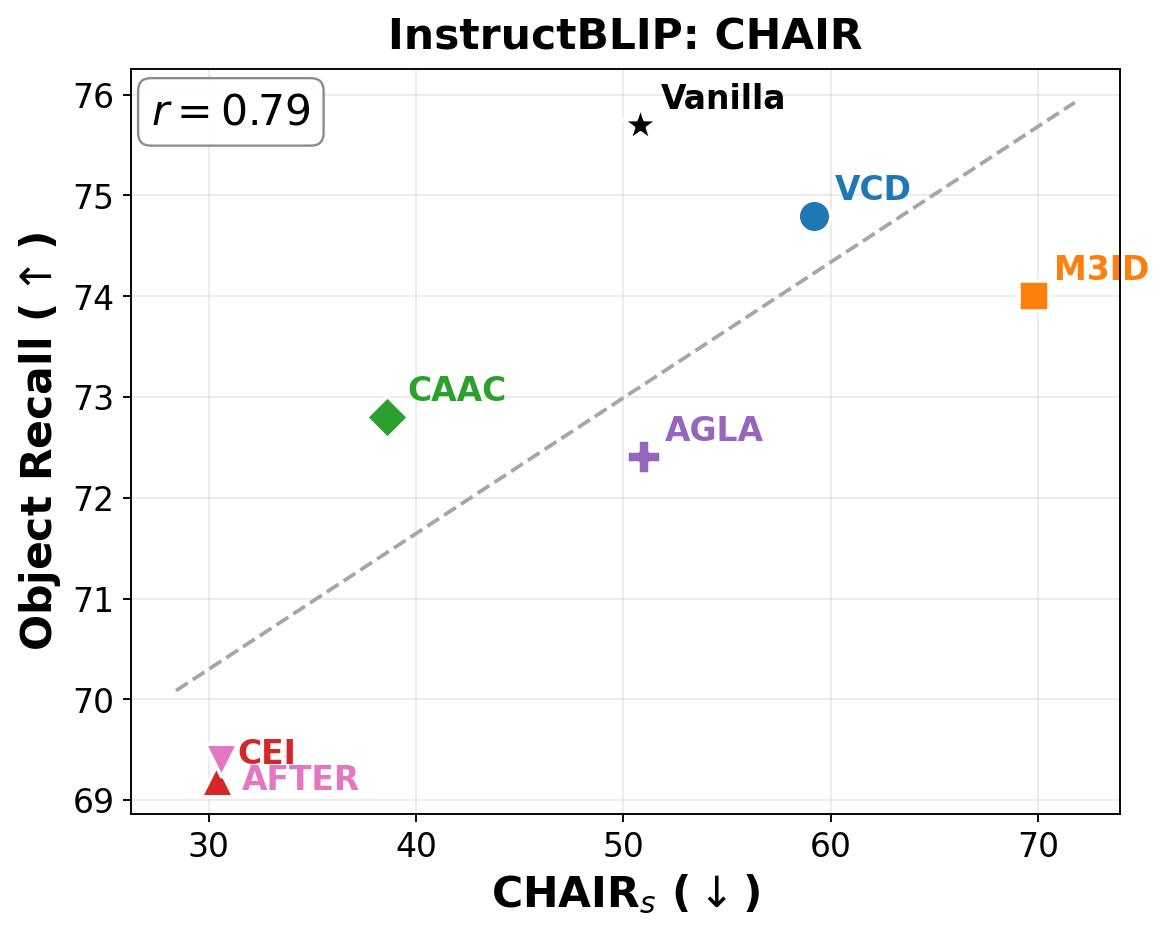}
    \caption{\chair: CHAIRs vs Recall}
    \label{fig:chair-instructblip}
\end{subfigure}
\begin{subfigure}{0.38\textwidth}
    \centering
    \includegraphics[width=\linewidth]{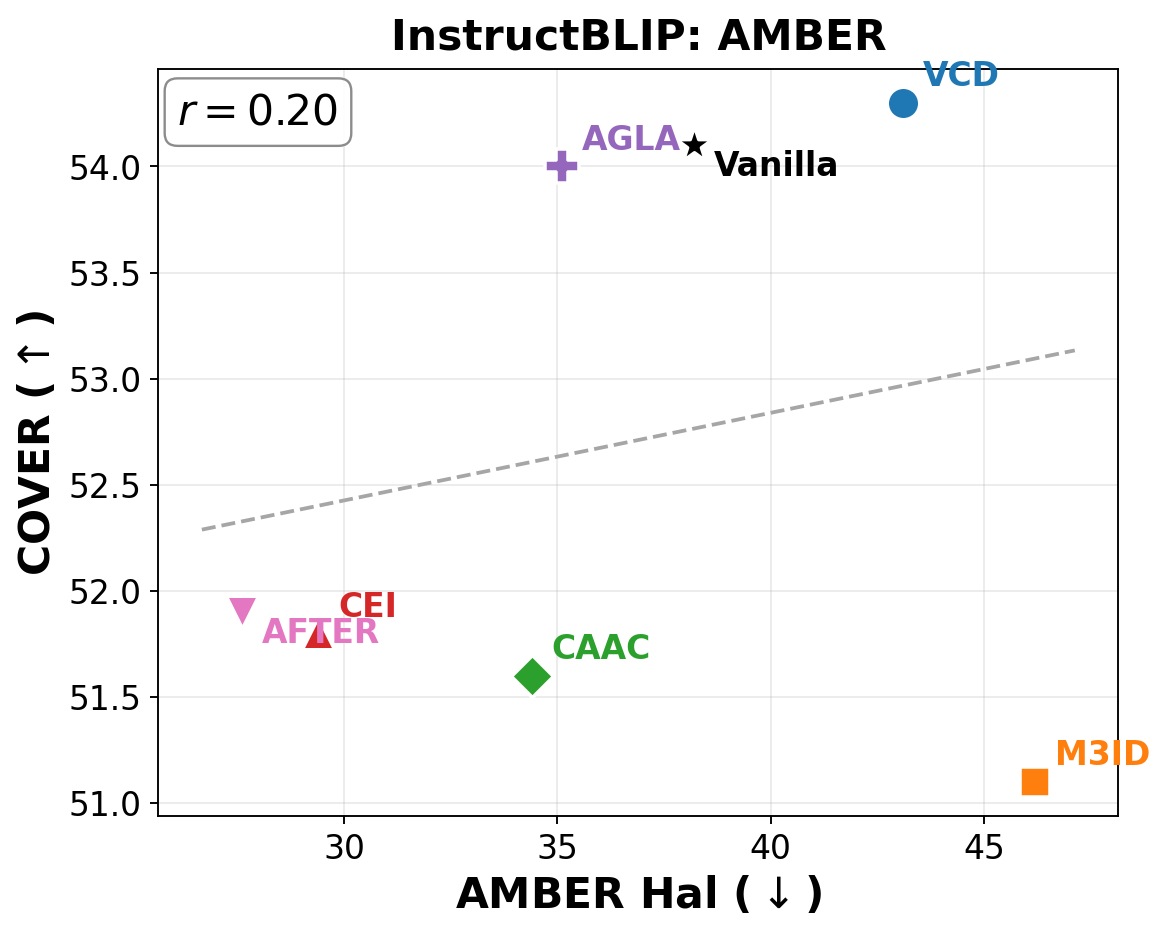}
    \caption{\amber: Hal vs Cover}
    \label{fig:amber-instructblip}
\end{subfigure}
\caption{Hallucination--informativeness trade-off on \instructblip.}
\label{fig:appendix-instructblip-scatter}
\end{figure}

Figures~\ref{fig:appendix-llava15-scatter} and \ref{fig:appendix-instructblip-scatter} display the decoupled hallucination--informativeness trade-off scatter plots for \llavaonefive and \instructblip, providing cross-architecture support for the faithfulness-informativeness trade-off. 
Across both additional foundation models: localized drops in hallucination error rates are fundamentally entangled with a loss of detailedness. Perhaps the only exception is \llavaonefive with \amber (Figure~\ref{fig:amber-llava15}), where AFTER manages to reduce hallucination while increasing the Cover.
On the other hand, on \chair, the positive association between hallucination error rate and object recall is exceptionally pronounced both for \llavaonefive (Figure~\ref{fig:chair-llava15}) and \instructblip (Figure~\ref{fig:chair-instructblip}). 
As illustrated in Figure~\ref{fig:appendix-llava15-scatter}, top-performing mitigation methods like CEI and CAAC minimize nominal error rates entirely via \textit{risk suppression}, forcing the model into the lowest-left margins where object recall sinks below 75\% and visual coverage settles under 49\%.

\subsection{CHAIR}

Table~\ref{tab:chair-full} reports the complete \chair results across the three LVLMs and seven decoding configurations (vanilla plus six mitigation methods). For each cell we report sentence-level hallucination ($\text{CHAIR}_s$, lower is better), instance-level hallucination ($\text{CHAIR}_i$, lower is better), object recall (higher is better), and the average response length in tokens. Crucially, a key observation from these results is that no single method is capable of simultaneously minimizing hallucination rates while maximizing object coverage. Instead, the configurations exhibit distinct trade-offs; for instance, contrastive decoding methods generally excel at maintaining high coverage, though this often comes at the cost of higher hallucination rates.

\begin{table}[t]
\centering
\small
\begin{tabular}{lcccc}
\toprule
Method & C$_s$ $\downarrow$ & C$_i$ $\downarrow$ & Recall $\uparrow$ & Len \\
\midrule
\llavanext       & 33.0 & 9.1  & 64.7 & 173.0 \\
\quad + VCD      & 42.6 & 10.7 & \textbf{69.9} & 170.6 \\
\quad + M3ID     & 40.8 & 12.3 & \underline{65.2} & 189.6 \\
\quad + CAAC     & 30.0 & \underline{8.7}  & 64.8 & 151.7 \\
\quad + CEI      & 30.6 & 9.8  & 64.5 & 140.3 \\
\quad + AGLA     & \underline{29.6} & \textbf{8.2}  & 59.4 & 173.2 \\
\quad + AFTER    & \textbf{25.8} & 9.3  & 62.9 & 216.3 \\
\midrule
\llavaonefive    & 46.6 & 14.8 & 79.6 & 88.1  \\
\quad + VCD      & 53.8 & 16.0 & \underline{80.0} & 102.0 \\
\quad + M3ID     & 56.2 & 15.7 & \textbf{82.2} & 94.4  \\
\quad + CAAC     & \underline{38.2} & 12.1 & 74.4 & 78.4  \\
\quad + CEI      & \textbf{31.0} & \textbf{9.5}  & 68.6 & 75.3  \\
\quad + AGLA     & 55.2 & 14.7 & 78.2 & 101.3 \\
\quad + AFTER    & 39.6 & \underline{10.5} & 76.5 & 96.6  \\
\midrule
\instructblip    & 50.8 & 15.9 & \textbf{75.7} & 102.1 \\
\quad + VCD      & 59.2 & 17.1 & \underline{74.8} & 104.1 \\
\quad + M3ID     & 69.8 & 21.3 & 74.0 & 102.0 \\
\quad + CAAC     & 38.6 & 11.6 & 72.8 & 94.0  \\
\quad + CEI      & \textbf{30.4} & \underline{8.0}  & 69.2 & 103.0 \\
\quad + AGLA     & 51.0 & 13.5 & 72.4 & 106.1 \\
\quad + AFTER    & \underline{30.6} & \textbf{7.6}  & 69.4 & 97.7  \\
\bottomrule
\end{tabular}
\caption{Full \chair results. Len denotes average response length in tokens. The best and second-best values for each performance metric are boldfaced and underscored respectively.}
\label{tab:chair-full}
\end{table}

\begin{table}[t]
\centering
\small
\begin{tabular}{lcccc}
\toprule
Method & CHAIR $\downarrow$ & Cover $\uparrow$ & Hal $\downarrow$ & Cog \\
\midrule
\llavanext       & 9.2  & \underline{61.1} & 50.3 & 5.0 \\
\quad + VCD      & 11.1 & \textbf{63.6} & 59.2 & 5.3 \\
\quad + M3ID     & 13.4 & 60.4 & 61.3 & 5.5 \\
\quad + CAAC     & \underline{8.6}  & 60.5 & \underline{48.2} & 4.7 \\
\quad + CEI      & 10.2 & 60.7 & 51.1 & 5.5 \\
\quad + AGLA     & 8.8  & 59.9 & 50.3 & 4.5 \\
\quad + AFTER    & \textbf{7.1}  & 58.7 & \textbf{35.6} & 2.4 \\
\midrule
\llavaonefive    & 7.4  & 48.7 & 31.3 & 3.4 \\
\quad + VCD      & 10.2 & 51.0 & 44.0 & 4.3 \\
\quad + M3ID     & 8.0  & \textbf{56.9} & 43.5 & 3.2 \\
\quad + CAAC     & \textbf{5.4}  & 48.3 & \textbf{23.0} & 2.1 \\
\quad + CEI      & \textbf{5.4}  & 47.3 & \underline{25.7} & 1.8 \\
\quad + AGLA     & 7.6  & 51.0 & 36.7 & 3.9 \\
\quad + AFTER    & \underline{6.4}  & \underline{51.4} & 30.3 & 3.6 \\
\midrule
\instructblip    & 8.4  & \underline{54.1} & 38.2 & 4.1 \\
\quad + VCD      & 9.3  & \textbf{54.3} & 43.1 & 4.3 \\
\quad + M3ID     & 10.3 & 51.1 & 46.2 & 4.7 \\
\quad + CAAC     & 7.7  & 51.6 & 34.4 & 3.5 \\
\quad + CEI      & \underline{6.3}  & 51.8 & \underline{29.4} & 2.6 \\
\quad + AGLA     & 7.5  & 54.0 & 35.1 & 3.8 \\
\quad + AFTER    & \textbf{5.7}  & 51.9 & \textbf{27.6} & 2.8 \\
\bottomrule
\end{tabular}
\caption{Full \amber results. Cog denotes the cognitive metric reported by AMBER. The best and second-best values for each performance metric are boldfaced and underscored respectively.}
\label{tab:amber-full}
\end{table}

\subsection{AMBER}

Table~\ref{tab:amber-full} reports the complete \amber results. For each (model, method) cell we report AMBER's $\text{CHAIR}$ (object-existence hallucination, lower is better), $\text{Cover}$ (fraction of ground-truth objects mentioned, higher is better), and $\text{Hal}$ (instance-level hallucination rate including attributes and relations, lower is better). Throughout the main paper we pair $\text{Hal}$ with $\text{Cover}$ as the hallucination--informativeness signal on AMBER. Mirroring the trade-offs observed on CHAIR, methods that score exceptionally well on reducing hallucination metrics ($\text{CHAIR}$ and $\text{Hal}$) often lag significantly behind on $\text{Cover}$. This consistent drop in coverage reinforces our primary claim that \textbf{most inference-time mitigation methods rely heavily on a risk-averse strategy—suppressing hallucinations by simply reducing visual commitment} rather than genuinely enhancing multimodal grounding.

\subsection{MMStar}

Table~\ref{tab:mmstar-full} reports per-category and overall accuracy on \mmstar for each (model, method) cell. The six category columns are Coarse Perception (CP), Fine-grained Perception (FP), Instance Reasoning (IR), Logical Reasoning (LR), Science \& Technology (ST), and Mathematics (MA); the final column is the macro average over the six categories. These absolute accuracies are the source data underlying the per-category change ($\Delta$ vs.\ vanilla) heatmap in Figure~\ref{fig:mmstar-delta}. Two critical patterns emerge from these absolute scores. First, no single mitigation method consistently performs well across all capability axes or across the three baseline LVLMs, with the granular results exhibiting high volatility. Second, the vanilla decoding configuration proves remarkably competitive, securing the second-highest overall accuracy once (on InstructBLIP) and the third-highest twice (on LLaVA-NeXT and LLaVA-1.5). Taken together, these trends demonstrate that current inference-time mitigation methods fail to reliably improve—and frequently degrade—the holistic multimodal capabilities of the base models.

\begin{table*}[t]
\centering
\small
\begin{tabular}{llccccccc}
\toprule
Model & Method & CP & FP & IR & LR & ST & MA & Overall \\
\midrule
\multirow{7}{*}{\llavanext}
  & Vanilla  & 0.568 & \underline{0.316} & 0.408 & 0.292 & 0.176 & 0.252 & 0.336 \\
  & VCD      & 0.560 & 0.304 & \textbf{0.436} & \textbf{0.308} & \underline{0.220} & \textbf{0.284} & \textbf{0.352} \\
  & M3ID     & 0.556 & 0.304 & 0.400 & \underline{0.296} & 0.216 & 0.236 & 0.335 \\
  & CAAC     & 0.536 & 0.288 & 0.400 & \underline{0.296} & 0.204 & \textbf{0.284} & 0.335 \\
  & CEI      & \underline{0.572} & \textbf{0.328} & 0.408 & 0.288 & \textbf{0.232} & 0.196 & \underline{0.337} \\
  & AGLA     & 0.544 & 0.256 & \underline{0.432} & 0.228 & \textbf{0.232} & \underline{0.280} & 0.329 \\
  & AFTER    & \textbf{0.580} & 0.288 & 0.424 & 0.260 & 0.192 & 0.240 & 0.331 \\
\midrule
\multirow{7}{*}{\llavaonefive}
  & Vanilla  & \underline{0.568} & \underline{0.272} & 0.388 & \underline{0.292} & \underline{0.252} & 0.196 & \underline{0.328} \\
  & VCD      & 0.556 & 0.240 & \underline{0.400} & \textbf{0.300} & 0.212 & 0.256 & 0.327 \\
  & M3ID     & 0.564 & \textbf{0.316} & 0.388 & 0.280 & 0.216 & 0.236 & \textbf{0.333} \\
  & CAAC     & 0.488 & 0.256 & 0.356 & \underline{0.292} & \textbf{0.280} & \textbf{0.276} & 0.325 \\
  & CEI      & \underline{0.568} & 0.236 & 0.376 & 0.264 & 0.228 & 0.212 & 0.314 \\
  & AGLA     & \underline{0.568} & 0.240 & 0.396 & 0.268 & 0.148 & \underline{0.260} & 0.313 \\
  & AFTER    & \textbf{0.572} & 0.248 & \textbf{0.408} & 0.276 & 0.232 & 0.208 & 0.324 \\
\midrule
\multirow{7}{*}{\instructblip}
  & Vanilla  & 0.340 & \underline{0.300} & 0.388 & 0.348 & 0.236 & \textbf{0.308} & 0.320 \\
  & VCD      & 0.316 & 0.296 & 0.296 & 0.292 & 0.256 & 0.256 & 0.285 \\
  & M3ID     & 0.360 & \textbf{0.304} & 0.316 & 0.324 & \textbf{0.264} & 0.280 & 0.308 \\
  & CAAC     & 0.340 & 0.268 & 0.356 & \underline{0.300} & 0.228 & 0.300 & 0.299 \\
  & CEI      & 0.348 & \textbf{0.304} & 0.344 & 0.232 & 0.228 & 0.240 & 0.283 \\
  & AGLA     & \textbf{0.420} & 0.296 & \textbf{0.408} & \textbf{0.380} & 0.232 & 0.292 & \textbf{0.338} \\
  & AFTER    & \underline{0.412} & 0.280 & \underline{0.404} & \underline{0.368} & \underline{0.260} & 0.264 & \underline{0.331} \\
\bottomrule
\end{tabular}
\caption{Full \mmstar results across the 3$\times$7 model$\times$method matrix.}
\label{tab:mmstar-full}
\end{table*}

\section{Qualitative Examples}
\label{app:qualitative}

\begin{figure*}[t]
\centering
\begin{subfigure}{0.49\textwidth}
    \centering
    \includegraphics[width=\linewidth]{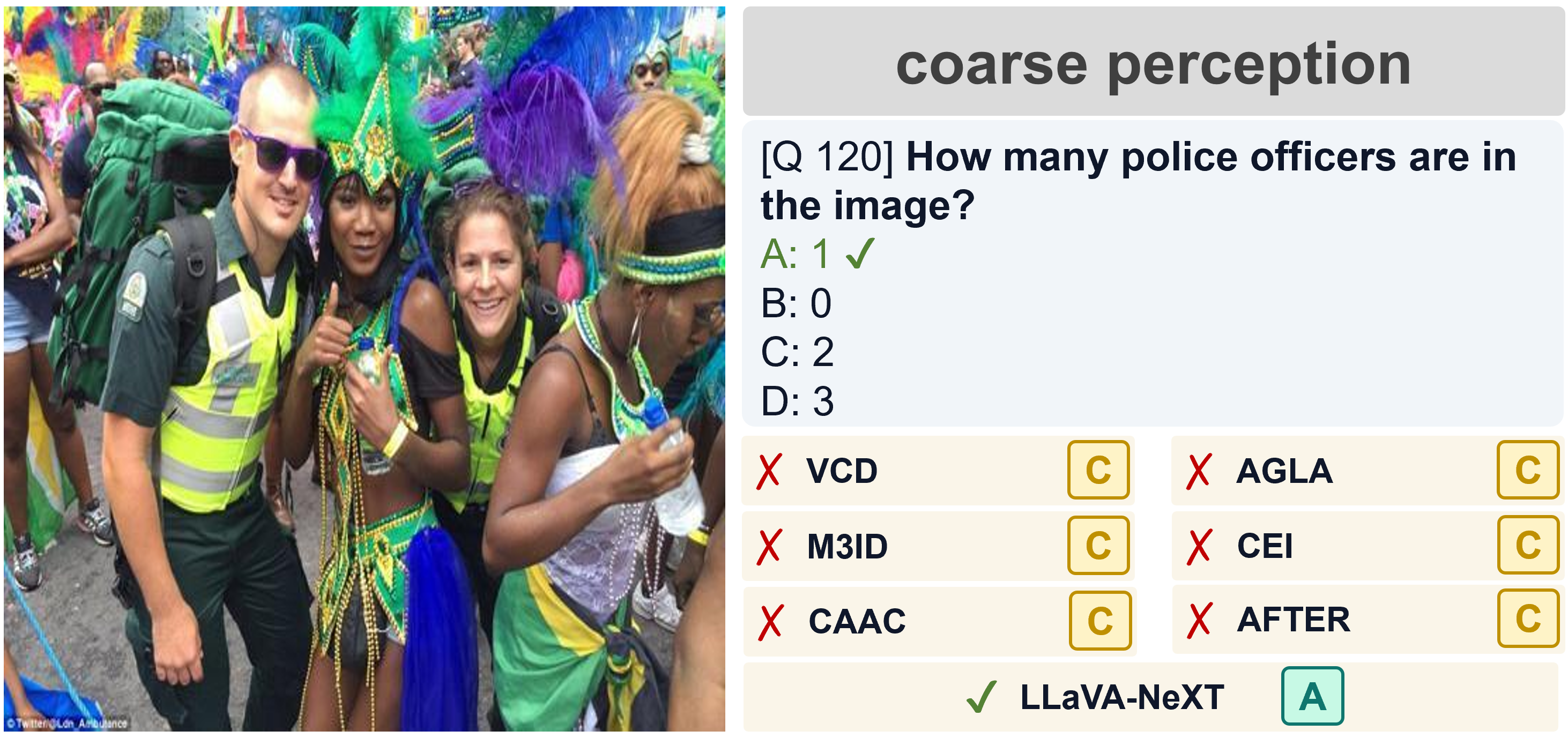}
    \caption{Coarse Perception Failure}
    \label{fig:case_police}
\end{subfigure}
\hfill
\begin{subfigure}{0.49\textwidth}
    \centering
    \includegraphics[width=\linewidth]{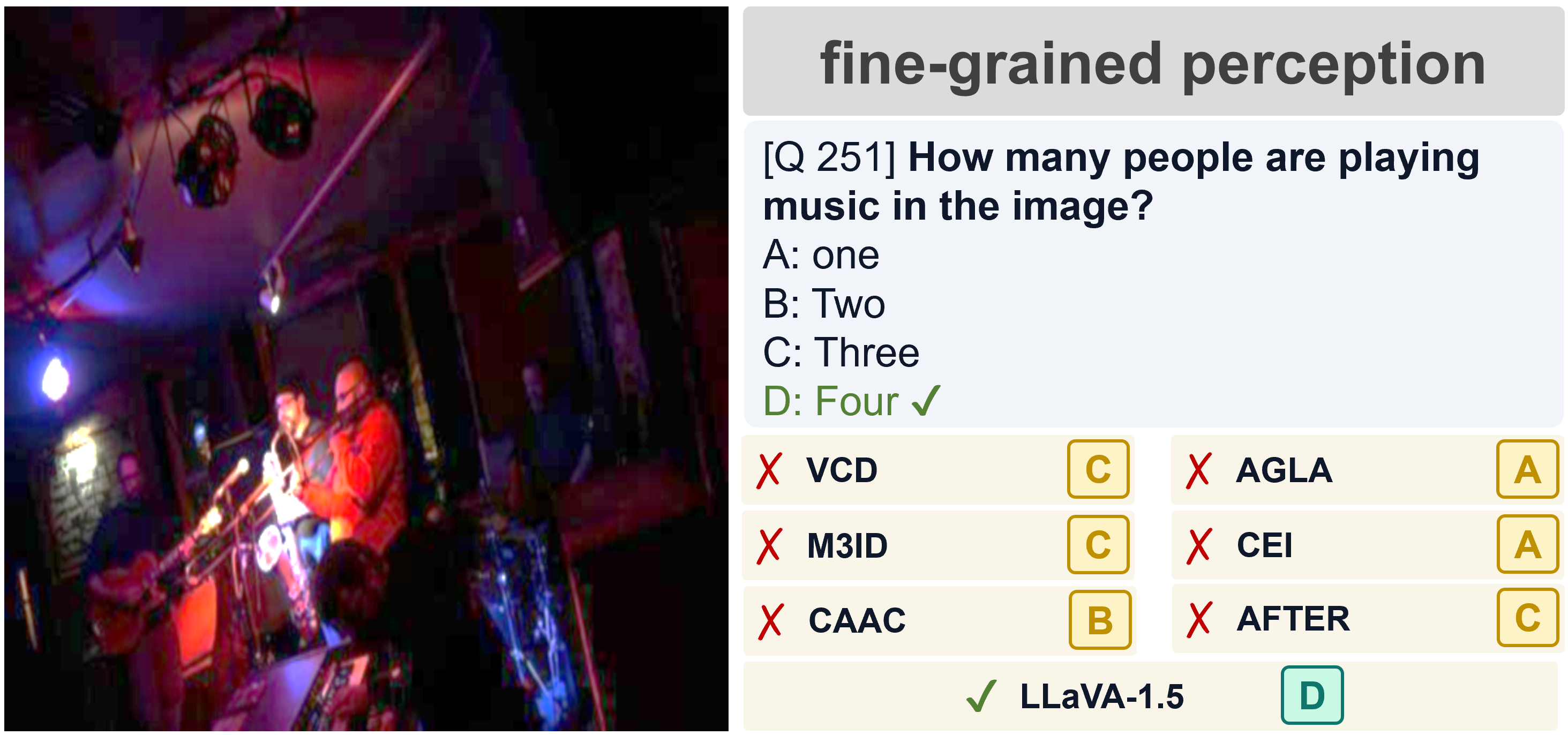}
    \caption{Fine-Grained Perception Failure}
    \label{fig:case_music}
\end{subfigure}

\vspace{4mm}

\begin{subfigure}{0.49\textwidth}
    \centering
    \includegraphics[width=\linewidth]{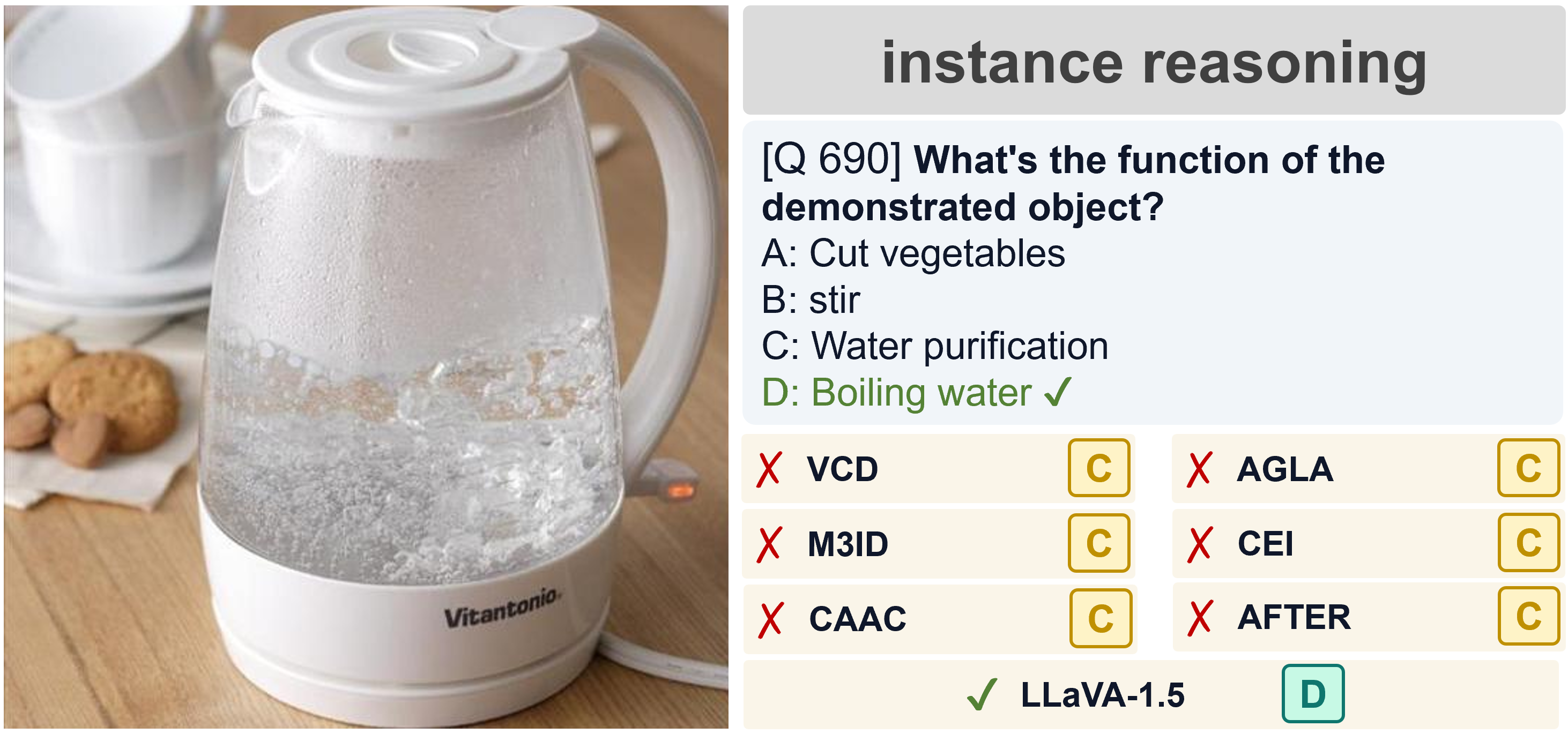}
    \caption{Instance Reasoning Failure}
    \label{fig:case_kettle}
\end{subfigure}
\hfill
\begin{subfigure}{0.49\textwidth}
    \centering
    \includegraphics[width=\linewidth]{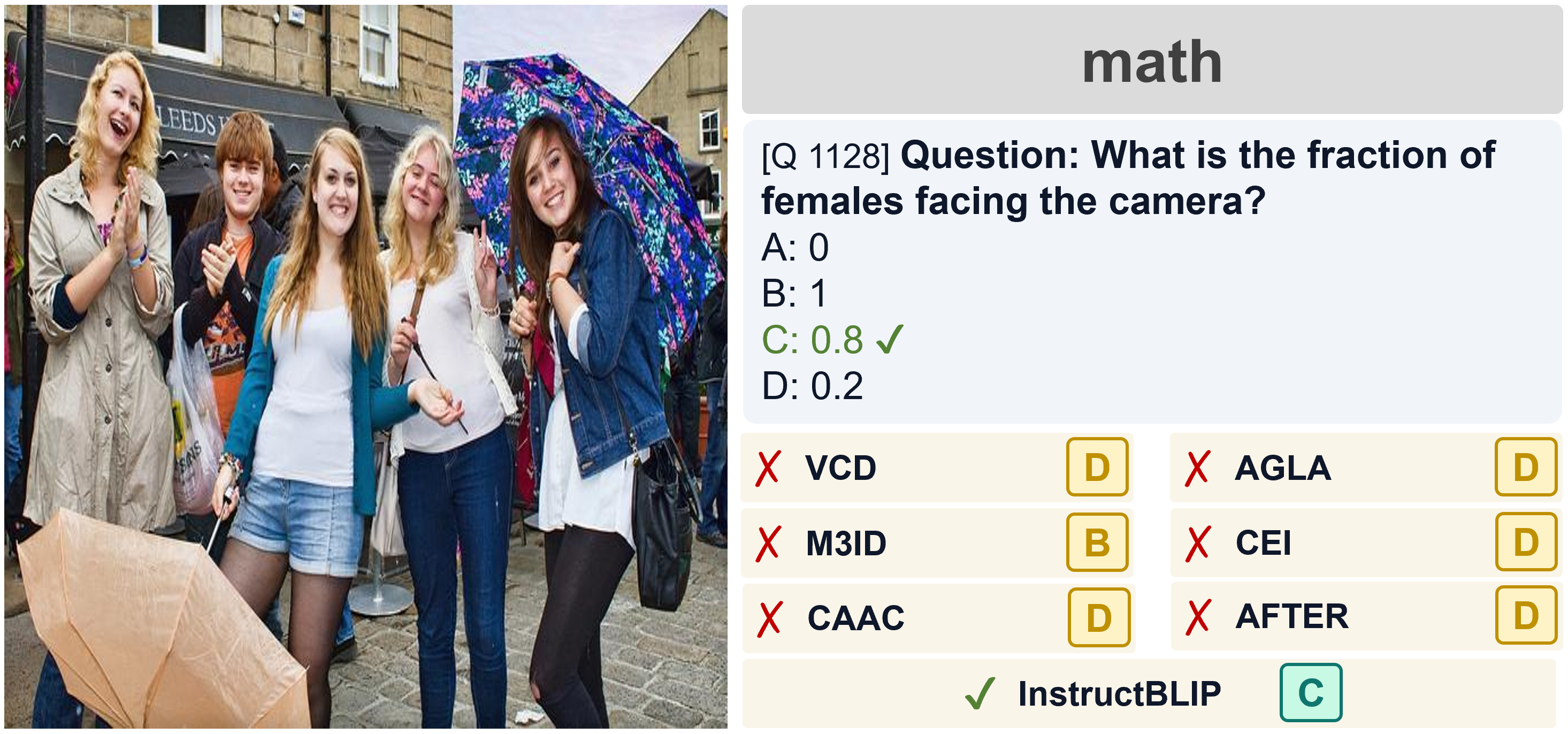}
    \caption{Mathematical Ratio Calculation Failure}
    \label{fig:case_math}
\end{subfigure}

\caption{Extended comparative case studies on \mmstar. Across perception, instance tracking, and fractional calculation, the base models' vanilla decoding output correct choices (green checkmarks). In contrast, the six inference-time mitigation strategies consistently induce systematic prediction failures across all evaluation tasks.}
\label{fig:appendix_grid}
\end{figure*}

To ground our macro findings, Figure~\ref{fig:appendix_grid} presents four verbatim task logs from \mmstar illustrating the transition from \textit{risk suppression} to \textit{capability cannibalization}. 
Across all core tasks, the unmitigated base models output natively correct choices (green checkmarks), whereas the mitigation methods consistently produce systematic errors. 
In Figure~\ref{fig:case_police}, the mitigation methods over-compensate for background density to incorrectly over-count crowd targets (\textbf{C}). 
Similarly, Figure~\ref{fig:case_music} demonstrates that \textit{risk suppression} alters confidence thresholds such that fine-grained details are filtered out, defaulting to understated group counts (\textbf{A} or \textbf{B}). 

This performance degradation is most acute during complex reasoning. 
In Figure~\ref{fig:case_kettle}, the mitigation methods squash vital semantic pathways, causing all six options to collapse onto a false generic attribute (\textbf{C}). 
Additionally, Figure~\ref{fig:case_math} shows that flattening the token distribution space breaks the fragile sequential token commitments vital for exact arithmetic, causing every evaluated framework to deviate into an incorrect ratio fraction (\textbf{D}). 

% --- FULL PAGE 2x2 GRID FIGURE ---

Two additional cases that show the failure mode of contrastive decoding methods are shown below where amplifying the visual-dependent tokens can force a wrong prediction.

% \subsection{Capability Regressions on \mmstar}

\begin{qualexample}[Confidently wrong on a math question]
\label{fig:qual1}
\centering
\includegraphics[width=0.55\linewidth]{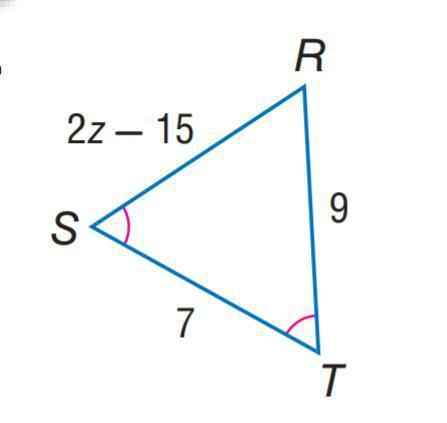}\par\vspace{6pt}
\raggedright
\qfield{Benchmark}{\mmstar}
\qfield{Model}{\llavanext}
\qfield{Method}{VCD}
\qfield{Category}{mathematics}
\qfield{Question}{Find $z$.}
\qfield{Options}{(A) 7~~(B) 12~~(C) 9~~(D) 15}
\qfield{Ground truth}{B}
\qfield{Model response}{\emph{``The correct answer is (D) 15.''}}
\qfield{Likely explanation}{Wrong (predicted D, correct B). The method commits confidently to a fluent but incorrect option which appears in the image. This could be due to amplifying the image-dependent tokens in VCD.}
\end{qualexample}

\begin{qualexample}[Wrong commitment on a food-web question]
\label{fig:qual2}
\centering
\includegraphics[width=0.6\linewidth]{figures/ex2_mmstar_llava15_m3id_idx773.png}\par\vspace{6pt}
\raggedright
\qfield{Benchmark}{\mmstar}
\qfield{Model}{\llavaonefive}
\qfield{Method}{M3ID}
\qfield{Category}{logical reasoning}
\qfield{Question}{According to the given food web, which is the animal that eats other animals and also becomes a prey to other animals?}
\qfield{Options}{(A) salmon~~(B) deer~~(C) moose~~(D) none of the above}
\qfield{Ground truth}{D}
\qfield{Model response}{\emph{``A: salmon.''}}
\qfield{Likely explanation}{Wrong (predicted A, correct D). The method commits to a specific incorrect option which appears in the image rather than reasoning about the question, a side effect of amplifying image-dependent tokens.}
\end{qualexample}

% \subsection{Side-by-Side: Same Image, Different Methods}

% \begin{qualexample}[Two methods regress on the same item]
% \centering
% \includegraphics[width=0.6\linewidth]{figures/ex5_mmstar_llavanext_idx1427.png}\par\vspace{6pt}
% \raggedright
% \qfield{Benchmark}{\mmstar}
% \qfield{Model}{\llavanext}
% \qfield{Category}{science \& technology}
% \qfield{Question}{Which label represents a reverse fault?}
% \qfield{Options}{(A) C~~(B) B~~(C) D~~(D) A}
% \qfield{Ground truth}{C}
% \qfield{Response (VCD)}{\emph{``C: D.''}}
% \qfield{Response (M3ID)}{\emph{``D.''}}
% \qfield{Likely explanation}{Both methods are wrong: VCD produces a confused label, while M3ID commits cleanly to the wrong option.}
% \end{qualexample}

% \subsection{Conservative Captioning (Candidate)}

% \begin{qualexample}[Example C.6 --- Caption omits salient content]
% \centering
% \includegraphics[width=0.6\linewidth]{figures/ex6_chair_llava15_imgid188824.jpg}\par\vspace{6pt}
% \raggedright
% \qfield{Benchmark}{\chair}
% \qfield{Model}{\llavaonefive}
% \qfield{Methods}{VCD and M3ID}
% \qfield{Image}{MSCOCO val2014, image\_id 188824}
% \qfield{Prompt}{Please describe this image in detail.}
% \qfield{Outcome (candidate)}{On this image, both VCD and M3ID on \llavaonefive produce captions that describe the central activity but omit secondary visual entities. A complete vanilla-vs-method side-by-side will be added once vanilla predictions are integrated from the broader team.}
% \end{qualexample}

% \section{Extended Qualitative Examples}
% \label{app:qualitative}

\end{document}